\documentclass{article}
\usepackage{spconf}

\usepackage{amsthm}

\usepackage{amsmath}
\usepackage{amsfonts}
\usepackage{amssymb}
\usepackage{graphicx}

\usepackage{amssymb}
\usepackage{tikz}
\usepackage{amsthm}
\usepackage{comment}

\theoremstyle{remark}
\newtheorem{remark}{Remark}

\usepackage{mathcomSTEv4}
\usepackage[boxed]{algorithm2e}

\usepackage{algorithmic}
\usepackage{caption}
\usepackage{subcaption}
\usepackage{adjustbox}

\usepackage[hyphens]{url}
\usepackage[hidelinks]{hyperref}
\hypersetup{breaklinks=true}
\usepackage{cite}

\begin{document}
	
	\title{
		An efficient label-free analyte detection algorithm \\ for time-resolved spectroscopy
	}
	%
	\twoauthors
	{Stefano Rini 
	}
	{ 
		Department of Electrical \& Computer Engineering  \\
		NCTU, Taiwan
	}
	{ Hirotsugu Hiramatsu  
	}
	{ 
		Department of Applied Chemistry\\
		NCTU, Taiwan 
	}
	
	
	\maketitle

	\begin{abstract}
		Time-resolved spectral techniques play an important analysis tool in many contexts, from physical chemistry to bio-medicine.
		Customarily,  the label-free detection of analytes is manually performed by experts through the aid of  classic dimensionality-reduction methods, such as Principal Component Analysis (PCA) and Non-negative Matrix Factorization (NMF).
		This fundamental reliance on expert analysis for unknown analyte detection severely hinders the applicability and the throughput of these such techniques. 
		%
		%
		For this reason, in this paper, we formulate this detection problem  as an unsupervised learning problem  and propose a novel machine learning algorithm for label-free analyte detection.
		To show the effectiveness of the proposed solution, we consider the problem of detecting the amino-acids in Liquid Chromatography coupled with Raman spectroscopy (LC-Raman).
		%
		%
		%
		%
	\end{abstract}
	\begin{keywords}
		Chemometrics; Time-resolved spectroscopy; Liquid chromatography; Raman spectroscopy.
	\end{keywords}

	\section{Introduction}
	\label{Sec:Intro}
	Time-resolved (a.k.a. on-line) spectral methods study the variation of the eigen-states of a system over time: UV/Visible, IR, Raman, and X-ray spectroscopy can all be applied to study the time evolution
	of a host of biological, chemical and physical processes.
	\footnote{In the following we use the term ``time-resolved'' to indicate that the spectral measurements are taken at successive time instances. In the literature, the term ``online'' is sometimes used to describe this scenario.. } 
	%
	%
	%
	Given the high-dimensionality of the data and the differences in acquisition scenarios, the analysis of time-resolved spectral signal has relied on the analysis by human experts through PCA or NMF. 
	More recently, deep learning methods have show incredible promise  in performing supervised learning for spectral data \cite{weng2017combining,ho2019rapid}.
	Motivated by the rapid advancement of modern machine learning techniques in the analysis of spectral data, we propose rather unsupervised  learning algorithm for time-varying spectral.
	%
	%
	%
	%
	The performance of this algorithm is validated for LC-Raman measurement of the elution of four amino-acid.

	\noindent
	{\bf Literature Review:} Combined techniques of separative and analytical methods (known as ``hyphenated techniques'') have benefited the chemical analysis of natural samples such as metabolites, herbal medicines, and various pollutants. \cite{madsen2010chemometrics,aboulwafa2019authentication}, among others. 
	Such combinations 
	consists, for example, of the Liquid Chromatography (LC) and Gas Chromatography (GC) for the separation of multiple chemical constituents and the spectrometers for Ultraviolet–Visible (UV) absorption , fluorometer (fluo), Mass Spectrometer (MS), etc. for the identification of eluates.
	%
	For instance, \cite{meyer2013practical} considers  the combination LC-MS for the separation of analytes.
	In the paper, we propose an algorithm for time-resolved spectral data and apply it to LC-Raman experimental data.
	More specifically, in these experiments we have combined a Raman spectrometer with the liquid chromatography (LC-Raman) using vertical flow method \cite{li2019vertical,lo2020missing}, that is the sample solution flows from a pinhole so as to produce a laminar flow column.
	%
	%
	%
	%
	For this experimental setting, several procedures have been developed to recover the information about the number, elution time, and spectral pattern of eluates, such as PCA \cite{jolliffe2016principal},  NMF \cite{luce2016using} , multivariate curve resolution-alternating least squares (MCR-ALS) \cite{de2014multivariate}, among others. 
	
	\noindent
	{\bf Contributions:} our contributions are summarized as follows:

	\noindent
	{$\bullet$ \underline{ Sec. \ref{sec:System Model} -- System Model:} }
	we provide general formulation for  label-free analyte detection in time-resolved spectral measurements. 
	This formulation encompasses LC-Raman  as a special case.
	%
	%
	%
	%
	%
	
	\noindent
	{$\bullet$ \underline{ Sec. \ref{sec:Dimensionality reduction approach} -- Dimensionality Reduction Approach:} We review the state-of-the-art approach for unknown analyte detection.
		This approach relies on the input from an expert in spectroscopy and spectral analysis to correctly combine the PCA components and reconstruct the analytes spectrum and elution patterns. 
		%
		
		\noindent
		{$\bullet$ \underline{ Sec. \ref{sec:Detection Algorithm} -- Novel Detection Algorithm:}}
		we propose a novel algorithm for unsupervised analyte detection which does not require any  expert input. 
		Our algorithm relies on fitting of the time-frequency spectral features, followed by $k$-means of the time parameters to cluster the unknown analytes. 
		%
		
		\noindent
		{$\bullet$ \underline{ Sec. \ref{sec:Experimental Results} -- Experimental Results:}} 
		we apply the proposed algorithm to the LC-Raman data recording the elution of four amino-acids  diluted in an H2O and acetonitrile solution.

		\section{System Model} 
		\label{sec:System Model} 
		Let us describe the time-resolved spectral signal at time $t$ and frequency $f$ as
		\ea{
			Y[t,f] = \sum_{i=1}^K \la_k[t] X_{i}[f]+ \lao[t]  S[f]+N[t,f], 
			\label{eq:in out}
		}
		with  
		\ean{
			t & \in  \{0,\De_t, \ldots , T \},  \quad   \  T \triangleq N \De_t, \\
			f & \in \{f_{\min},f_{\min}+\De_f, \ldots, f_{\max} \},  \quad  f_{\max} \triangleq f_{\min}+M \De_f,
		}
		where $\De_t$/$\De_f$ is the time/frequency precision.
		%
		%
		In \eqref{eq:in out}, $K$ represent the number of analytes in the solution: the $i^{\rm th}$
		analyte  has a spectrum $X_k[f]$  and elution pattern $\la_k[t]$
		The solvent has spectrum $S[f]$ and elution pattern $\nu(t)$. 
		The set of $N \times M$ measurement in \eqref{eq:in out} is also corrupted by an additive i.i.d. noise $N[t,f]$ which is assumed to be standard normal distributed. 
		We further assume that the spectrum  of the $i^{\rm th}$ analyte is obtained as
		\ea{
			X_k[f]=\sum_{j=1 }^{N_k} L_{kj}[f] ,   \quad L_{kj}[f] \triangleq L[f,c_{kj},A_{kj},\Ga_{kj}],
			\label{eq X k f}
		}
		where $N_k$ is the number of peaks: the $j^{\rm th}$ peak is described by the location parameter, $c_{kj}$, the scale parameter, $A_{kj}$,  and the shape parameter  $\Ga_{kj}$.
		%
		The elution pattern described as 
		\ea{
			\la_k[t]=
			q_k G_k[t],  \quad G_k[t] \triangleq G[t,o_{k},d_k,\al_k,\be_k],
			\label{eq:la t}
		}
		where $q_k$ is the quantity of the $k^{\rm th}$ analyte,  $G_k[t]$  is the elution peak where $o_k$ is origin,  $d_k$ the duration, $\al_k$ the rise time and $\be_k$  the fall time.
		In \eqref{eq:la t} we assume that that the effect of quantity of the analyte on the elution pattern is linear; also, we assume that analyte elution patterns do not superpose. 
		\footnote{These assumption is well justified  when (i) the analyte concentration is orders of magnitude smaller than that of the solvent and (ii) analytes  . }
		The model in \eqref{eq:in out} is expressed as in matrix form as 
		\ea{
			\Yv= \La\Xv^T + \Lao \Sv^T +\Nv,
			\label{eq: vector form}
		} 
		where $\Yv,\Nv \in \Rbb_{+}^{N \times M}$,  $\La \in \Rbb_{+}^{N\times K}$, $\Xv \in \Rbb_{+}^{M \times K}$,  $\Lao \in \Rbb_{+}^{N}$, and  $\Sv \in \Rbb_{+}^{M}$, where $\Rbb_+$ is the set of positive real numbers.
		
		\noindent
		{\bf Unknown analyte detection problem:} 
		we consider here the unsupervised problem of detecting the presence of unknown analytes as described by their spectral peaks and elution patterns. 
		Let us denote the estimation outcome set  $\{\Xh_k[f],\lah_k[t]\}_{k=1}^{\Kh}$, where the hat indicates (here and also for the parameters in \eqref{eq X k f}, \eqref{eq:la t}, and \eqref{eq: vector form}) the outcome of the learning algorithm. 
		%
		%
		%
		
		Various metrics can be devised for evaluating the performance of the detection algorithm: for instance one can evaluate the precision in recovering the highest peak in a spectrum or the elution peak, in distinguishing analytes in a specific family or in recovering the support of the spectrum. 
		For simplicity, in the following we evaluate the performance as 
		%
		%
		\ea{
			\rho(\qv)= \f 1 {T (f_{\max}-f_{\min})} \sum_{i=1}^M  \f {\langle \Yv_i, \Yhv_i \rangle}{\| \Yv_i \|_2 \|\Yhv_i \|_2},
			\label{eq:projection}
		}
		where
		$\Yv_i$ indicates the $i^{\rm}$ row  of the matrix $\Yv$ and $\qv=[q_1,\ldots, q_k]$ is the vector containing the concentration of the analytes. 
		In \eqref{eq:projection} we have explicitly expressed the dependency of the detection performance on the analytes quantities.
		We do so in order to define the Limit Of Detection (LOD) of a particular experiment.
		Let $\cv=[q_1,\ldots,q_K]$ with $\| \cv\|=1$  containing the relative concentration of the $K$ analytes, then LOD for the set of relative concentration $\cv$  is defined as the smallest scalar $\eta$ such that 
		$\rho(\eta \cv)>1$, since the variance of the additive noise is larger than one. 
		\footnote{A more precise discussion of the LOD is deferred to \cite{ICASSPhh}.}

		\noindent
		{\bf LC-Raman analysis of amino-acids:} as an recurring example of the signal in Sec. \ref{eq:in out}, we consider the setting in Fig. \ref{fig:Raman}.
		The four amino-acids, namely glycine (gly -- blue), leucine (leu -- orange ), phenylalanine (phe -- green), and tryptophan (trp -- purple), in an injected  1mL mixture solution is conveyed with a mobile phase solution (light green) and applied 
		%
		to an high-performance LC (HPLC) column and attached to a hydrophobic resin.  The color coding in Fig. \ref{fig:amino} and is consistent through the paper.
		Each amino-acid is separately eluted from the column by gradually increasing the ratio of a hydrophobic, i.e. acetonitrile (ACE), to aqueous component (H2O) in the mobile phase using a programmable pump.  
		%
		%
		Successively this solution is analyzed through LC-Raman using vertical flow method, as described in \cite{li2019vertical}.
		The presence of the eluates can be detected in Fig. \ref{fig:Raman} as weak 2D-peaks rising over the spectrum of the solvent.  
		\begin{figure}
			\centering
			\includegraphics[width=\linewidth]{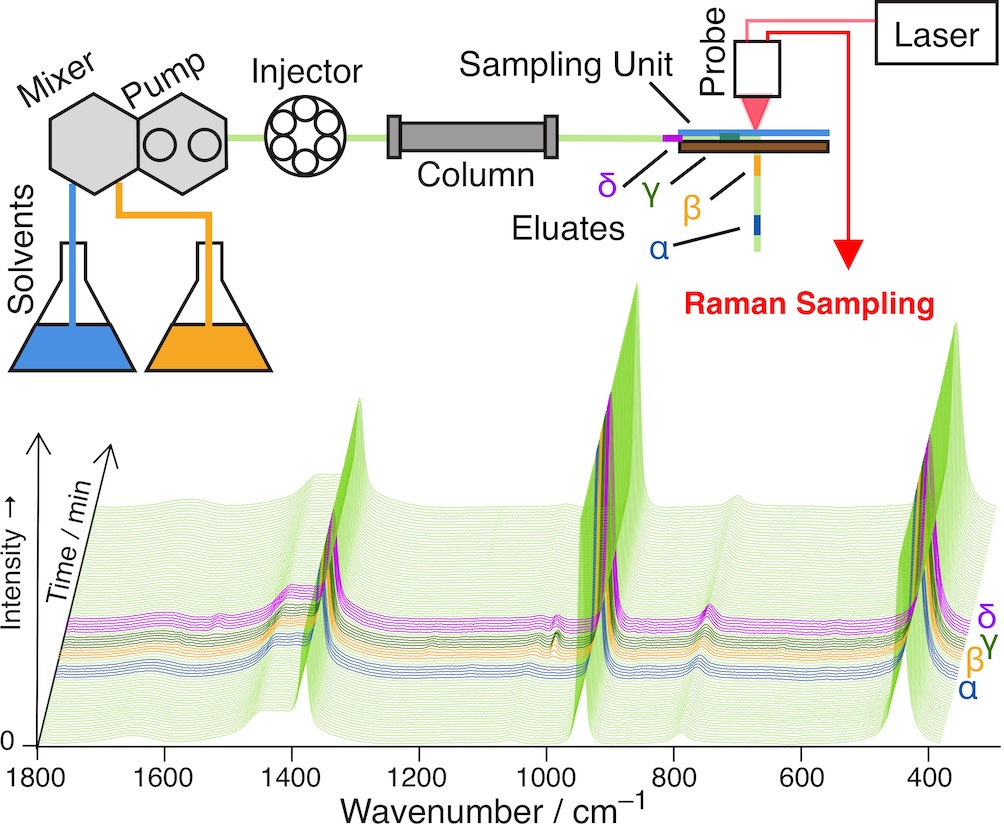}
			\vspace{-0.2 cm}
			\caption{The LC-Raman experiment set-up in Sec. \ref{sec:System Model}.}
			\vspace{-0.6 cm}
			\label{fig:Raman}
		\end{figure}
		%
		%
		The spectra of these amino-acids and their elution time are presented in Sec. \ref{fig:amino}.
		For this experimental setting, the difficult in detection of the analyte arises from (i) the small analyte concentration, in the  order of Millimolar (mM), and the fast elution times, in the order of minutes. 

		\begin{figure}
			\centering
			\begin{subfigure}{0.22\textwidth}
				\centering
				\includegraphics[trim={2cm 0 1.5cm 1.25cm},clip,width=\textwidth]{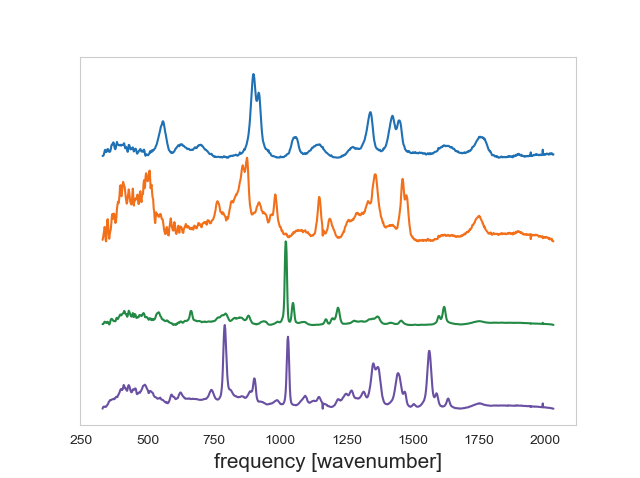}
				\caption{Frequency spectra}
				\label{fig:spectra}
			\end{subfigure}
			\hfill
			\begin{subfigure}{0.22\textwidth}
				\centering
				\includegraphics[trim={2cm 0 1.5cm 1.25cm},clip,width=\textwidth]{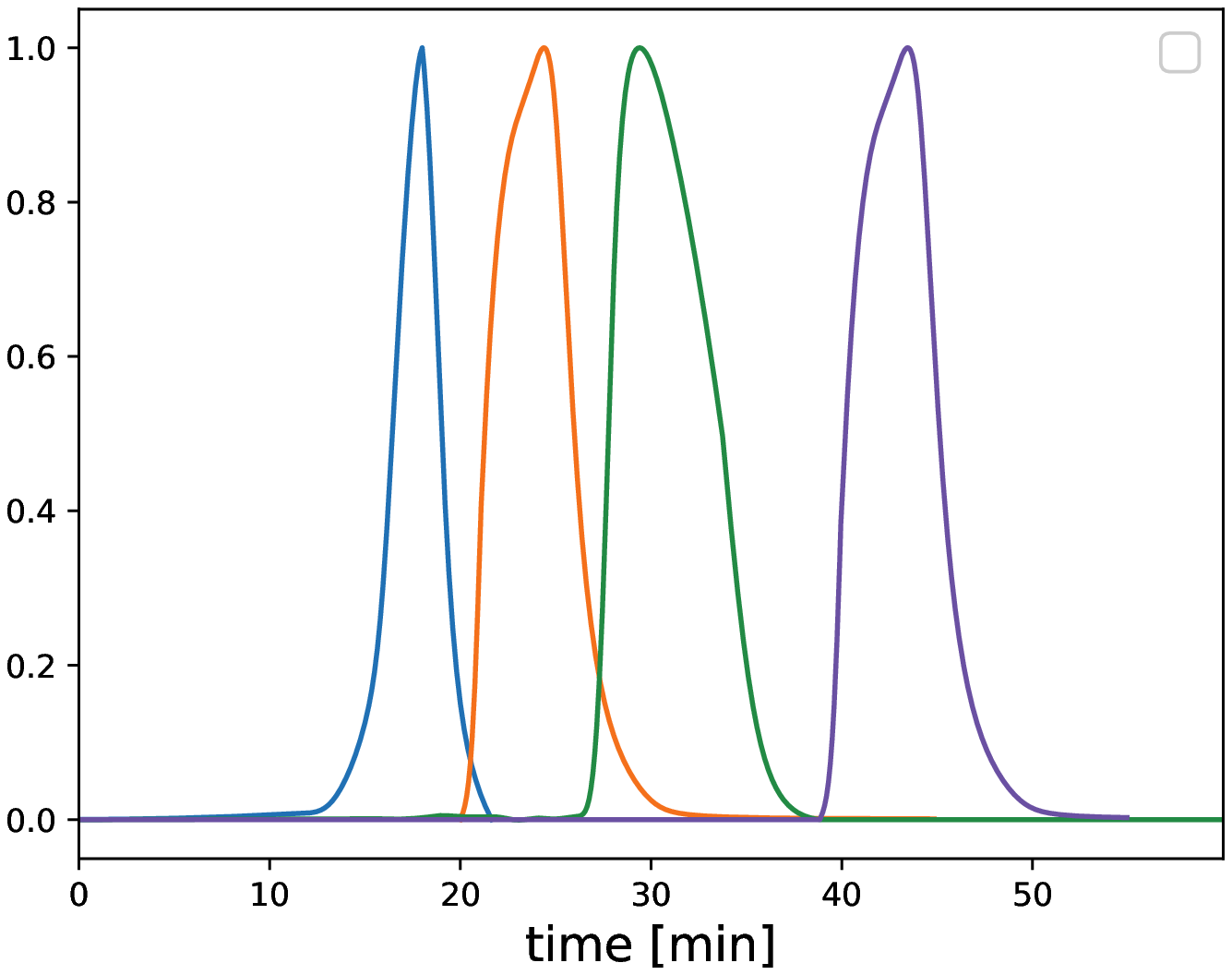}
				\caption{Time elution}
				\label{fig:time}
			\end{subfigure}
			\vspace{-0.1 cm}
			\caption{LC-Raman spectrum of four amino-acids:  glycine (gly -- blue), leucine (leu -- orange ), phenylalanine (phe -- green), and tryptophan (trp -- purple).}
			\label{fig:amino}
		\end{figure}
		\begin{figure}
			\centering
			\includegraphics[width=\linewidth]{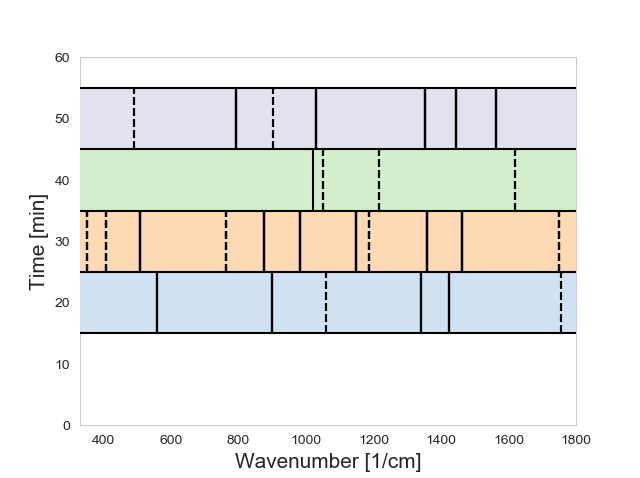}
			\vspace{-0.15 cm}
			\caption{Conceptual LOD plot for $\rho(\qv)$ it \eqref{eq:projection}.  }
			\vspace{-0.5 cm}
			\label{fig:detection plot}
		\end{figure}
		For this experimental setting, the detection performance as in \eqref{eq:projection} can be visualized as in Fig. \ref{fig:detection plot}.
		The horizontal bands represent the elution peaks of each analyte as in Fig. \ref{fig:time}. The vertical lines in each band represents the peaks in the spectra as in Fig. \ref{fig:spectra}: the solid line corresponds to the peaks with highest prominence (50\% percentile) while dotted lines correspond to peaks with average prominence (30\% percentile).
		By plotting the estimated parameters describing the spectrum and elution peaks in \eqref{eq X k f} and \eqref{eq:la t} in Fig. \ref{fig:detection plot}, one can visually estimate the portion of time/frequency features that are correctly estimated by the detection algorithm for a given analyte concentration $\qv$.
		
		\begin{remark}
			\label{rem:modelling}
			For brevity, we shall not discuss some important aspects of LC-Raman spectroscopy that further hinder the analyte detection tasks.
			%
			Such considerations are relegate to  App. \ref{app:modelling}
			
		\end{remark}
		
		\section{Dimensionality Reduction Approach}
		\label{sec:Dimensionality reduction approach}
		%
		Generally speaking, the state-of-the-art analysis of time-resolved spectral data relies on the input from a human expert as, so far, no effective algorithm as been devised for this task. 
		%
		The starting point for data analysis is usually a classic dimensionality reduction techniques, such as PCA or NMF. 
		%
		For the case of PCA, the  $\Yv$ in \eqref{eq: vector form} is approximated as $\Yv \approx \Uv \Vv^T$
		for some $J$ $\Uv \in \Rbb^{N \times \Kh}$, and $\Vv^T \in \Rbb^{\Kh \times M}$ for some $\Kh$ appropriately chosen. 
		%
		%
		%
		The output of the PCA is used to estimate an analyte time and frequency characteristics as  $\Lahv  = \Uv  \Tv$, $\Xhv  =\Vv  \Tv^{-1}$ 
		for some matrix $\Tv \in \Rbb^{\Kh \times \Kh}$ to be  carefully chosen by the expert.
		It is clear that this approach is substantially a trial-and-error approach in which the human expert relies on his/her own understanding of the physical phenomena underlying the data.
		As an example, let us consider Fig. \ref{fig:tuned PCA}.
		%
		%
		%
		%
		In Fig. \ref{fig:pca_f}/Fig. \ref{fig:pca_t}  the PCA output for frequency/time is shown: after the appropriate choice of $\Tv$ above, the spectra and elution windows as in Fig. \ref{fig:amino} are reconstructed as in 
		Fig. \ref{fig:adj_f}/Fig. \ref{fig:adj_t}.
		\begin{figure}
			\centering
			\begin{subfigure}{0.22\textwidth}
				\centering
				\includegraphics[trim={2cm 0 1.5cm 1cm},clip,width=\textwidth]{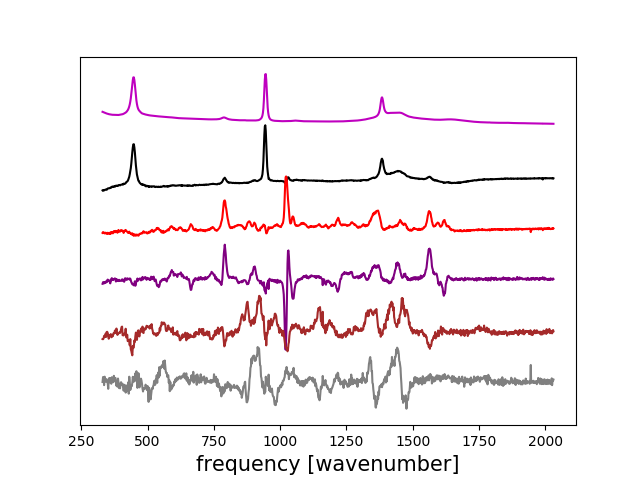}
				\caption{PCA--frequency}
				\label{fig:pca_f}
			\end{subfigure}
			\hfill
			\begin{subfigure}{0.22\textwidth}
				\centering
				\includegraphics[trim={2cm 0 1.5cm 1cm},clip,width=\textwidth]{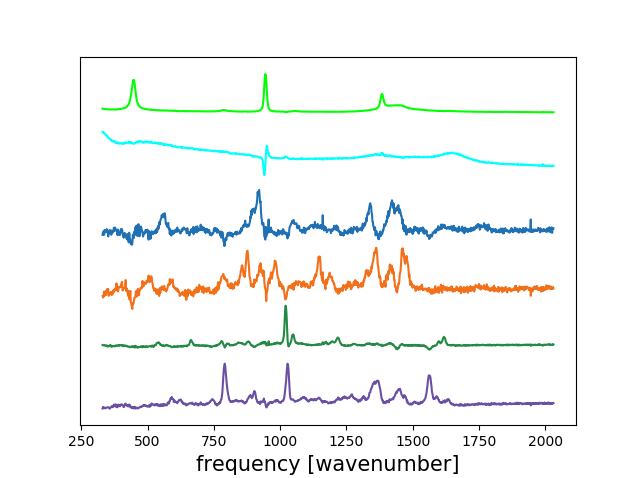}
				\caption{Tuned PCA--frequency}
				\label{fig:adj_f}
			\end{subfigure}
			\hfill
			\begin{subfigure}{0.22\textwidth}
				\centering
				\includegraphics[trim={2cm 0 1.5cm 1cm},clip,width=\textwidth]{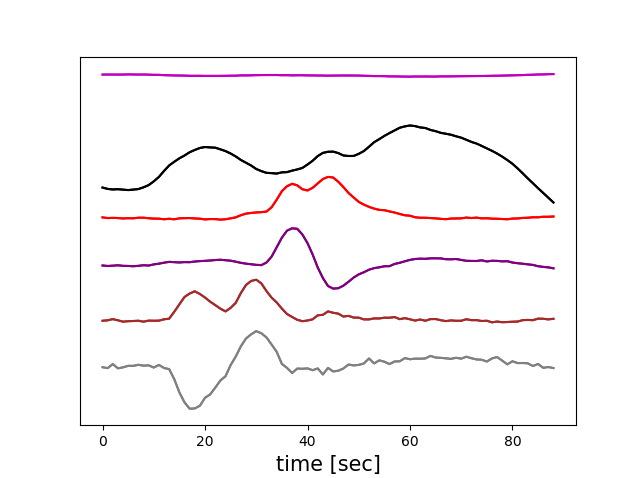}
				\caption{PCA--time}
				\label{fig:pca_t}
			\end{subfigure}
			\hfill
			\begin{subfigure}{0.22\textwidth}
				\centering
				\includegraphics[trim={2cm 0 1.5cm 1cm},clip,width=\textwidth]{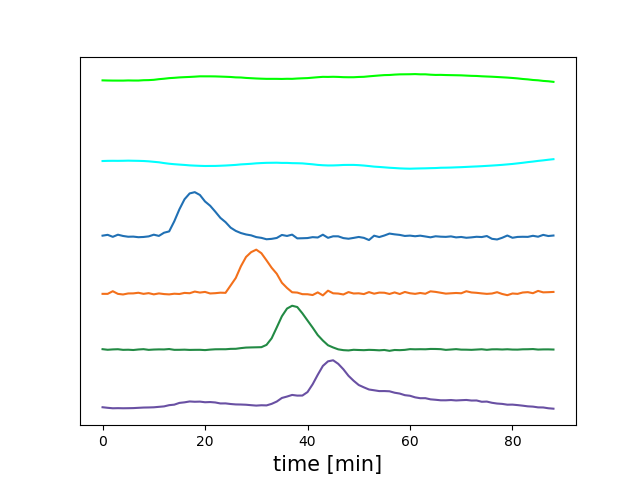}
				\caption{Tuned PCA--time}
				\label{fig:adj_t}
			\end{subfigure}
			\hfill
			\vspace{-0.15 cm}
			\caption{Tuning  of PCA components by a human expert.}
			\vspace{-0.25 cm}
			\label{fig:tuned PCA}
		\end{figure}
		%
		We note that the PCA components give to clear indication of the analyte features. Also note that one of the reconstructed components (cyan) is not associated with any analyte or solvent. 
		%
		%
		%
		%
		%
		%
		%
		%
		%
		\section{A Novel Detection Algorithm}
		\label{sec:Detection Algorithm}
		In this section, we propose a novel algorithm to automate the unknown analyte identification which does not rely on the aid from a human expert as for the approach in Sec. \ref{sec:Dimensionality reduction approach}.
		%
		%
		The algorithm is designed under the assumption that the spectral peaks and the time elution peaks can be well-modeled as in \eqref{eq X k f} and \eqref{eq:la t}.
		%
		%
		The pseudo-code of the proposed algorithm can be found in Alg. \ref{algo:detection}.
		%
		\begin{algorithm}
			\begin{algorithmic}[1]
				\STATE  {\bf Input:} { Measurements matrix $\Yv$, sensitivity level $\gamma$, fitting functions $L[f],G[t]$, solvent spectrum $\Sv$}
				\STATE  {\bf Output:}  Estimated analyte spectral/frequency \\  features $\Xhv$,$\Lahv$
				\STATE	Remove the solvent spectrum $\Sv$ from $\Yv$  through \\ linear regression over each time instant
				\STATE	Smooth the data using a Savitzky-Golay filter
				\STATE	Detect 2D peaks in $\Yv$ above the sensitivity level $\ga$
				\FOR{all peaks} 
				\STATE	fit $L[f] G[t]$ to the  $i$-th peak over $\Yv$, obtain \\  estimate parameters $\{ \Ah_i, \Gah_{i}, \ch_i , \widehat{o}_i, \widehat{d}_i, \alh_i, \beh_i \}$ 
				\ENDFOR
				%
				\STATE 	Cluster the parameters over the set $\{\widehat{o}_i, \widehat{d}_i \}$ using $K$-means algorithm. 
				Let  $\Kh$ be the number of \\ predicted  clusters 
				\STATE 	Label each peak  according to the $K$- meas output \\ to obtain $\Xhv$,$\Lahv$
				%
				%
			\end{algorithmic}
			\vspace{0.25 cm}
			\caption{Label-free analyte detection algorithm in Sec. \ref{sec:Detection Algorithm}.}
			\vspace{0.35 cm}
			\label{algo:detection}
			\vspace{-0.75cm}
		\end{algorithm}
		%
		%
		%
		%
		%
		%
		%
		%
		%
		
		\noindent
		{\bf LC-Raman analysis of amino-acids:} in this recurring scenario, we choose the the spectral peaks are modeled as pseudo-Voig peak, i.e.
		{
			\ea{
				L[f,c,A,\Ga] & =\nu \lb \f 1 {\sqrt{2\pi\sgs}} e^{-\f {(x-c)^2}{2 \sgs}} \rb  \label{eq:pseudo} \\ 
				& \quad \quad + (1-\nu) \f { \Gamma}{ \pi \lb (f-c)^2 + \Gamma^2 \rb}, \nonumber
				%
				%
			}
		}
		with $\nu \in [0,1]$ and for $\Gamma=\sqrt{2\log(2)\sgs}$, while the temporal peaks through a window with a trigonometric rise and fall transients, i.e.
		{
			\footnotesize 
			\ea{
				& G[t,M,o,d,\al,\be]=  
				\label{eq:cos win} \\
				& \lcb \p{
					\f M 2  \lb 1-\cos\lb \f {2 \pi (t-o)}{2 \al}\rb\rb & o < t \leq t+\al\\
					M &  o+\al < t \leq o+\al+d \\
					\f M 2 \lb 1-\cos\lb \f {2 \pi  (\be-(t-d-o-\al))}  {2 \be} \rb \rb     & o+\al+d < t  \\ 
					& \quad { \rm and }  \ \ t \leq o+\al+d+\be \\
					0  & {\rm otherwise}.
				}
				\rnone \nonumber 
			}
		}
		%
		%
		\section{Experimental Results}
		\label{sec:Experimental Results}
		In this section we apply the algorithm in Sec. \ref{sec:Detection Algorithm} to the experimental setting in which  $100$ mM of gly, $100$ mM of leu, $55$ mM of phe, and $55$ mM of trp  are eluted through LC Raman spectroscopy, as described in the previous sections.
		The frequency and time features are those presented in Sec. \ref{fig:amino}. The LOD  detection plot is that in Fig. \ref{fig:detection plot}. The exert analysis is that presented in Fig. \ref{fig:expert}.
		Here the intensity of the color corresponds to the height of the corresponding peak in the reconstructed spectrum. Note that, for clarity, we do not attempt to depict the other coefficients in 
		%
		%
		The results of the algorithm are presented in Fig. \ref{fig:us}.

		\begin{figure}[h]
			\centering
			\begin{subfigure}{0.45\textwidth}
				\centering
				\includegraphics[trim={2cm 0 1.5cm 1.25cm},clip,width=0.85\textwidth]{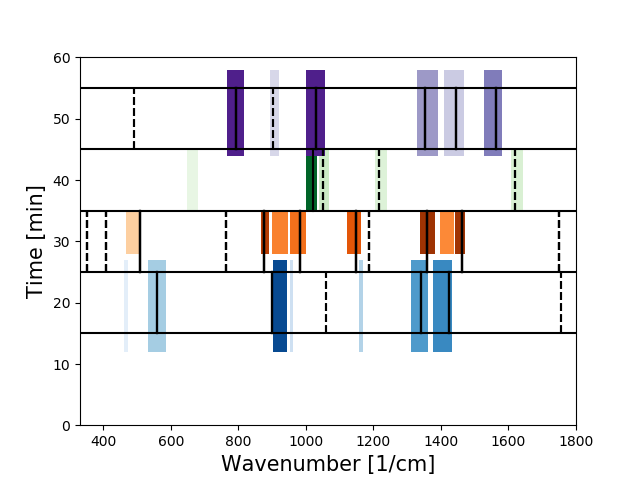}
				\caption{Expert detection performance.}
				\label{fig:expert}
			\end{subfigure}
			\hfill
			\begin{subfigure}{0.5\textwidth}
				\centering
				\includegraphics[trim={2cm 0 1.5cm 1.25cm},clip,width=0.85\textwidth]{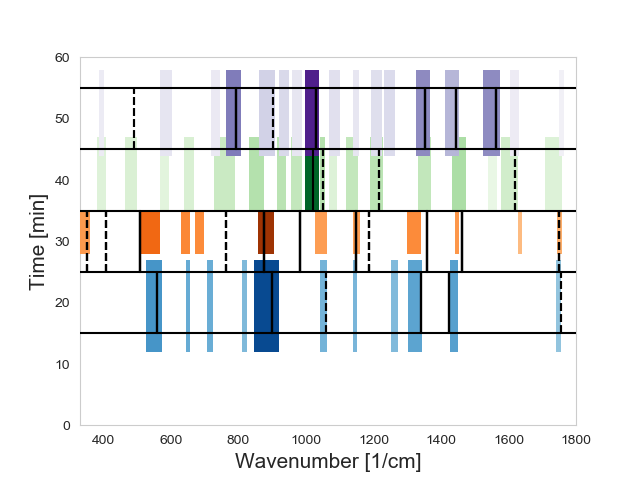}
				\caption{Proposed algorithm  detection performance.}
				\label{fig:us}
			\end{subfigure}
			\vspace{-0.15 cm}
			\caption{Comparison of the detection performance for the methods in Sec. \ref{sec:Dimensionality reduction approach} and in Sec. \ref{sec:Detection Algorithm}.}
			\label{fig:compare}
		\end{figure}

		\section{Conclusion}
		In this paper, a new unsupervised algorithm for unknown analyte detection has been proposed for time-resolved spectral measurements.
		This algorithm relies on the explicit modeling of the spectral and temporal peaks, together with a clustering algorithm applied over the temporal parameters.
		The effectiveness of the algorithm has been investigated for data collected using liquid-chromatography Raman spectroscopy applied to the elution of four amino-acids.
		
		\medskip
		
		\noindent
		{\bf Paper Reproducibility:}
		The experimental data in Sec. \ref{sec:Experimental Results} is provided  in \cite{dataset}.
		The authors thank Mr. Yuhao Lo  and Mr. Sergio Alejandro Diaz for the help in preparing such data.
		
		\newpage
		\bibliographystyle{vancouver}
		\bibliography{cyto_flow_bib}
		\newpage

\newpage
\onecolumn
\appendix

\section{Comments on Remark  1}
\label{app:modelling}

A more correct modeling of a Raman spectral measurement is as follows:
%
%
\ea{
Y[t,f] = \sum_{i=1}^K \la_k[t] \lb X_{i}[f]+X_{i ,{\rm FL}}[f] \rb + \lao[t] S[f]+N[t,f]+Z_{\rm SN}[t,f]+Z_{\rm CN}[t,f], 
\label{eq:in out v1}
}
The time signals $\la_k[t]$,$\lao[t]$, the frequency signals $X_{i}[f]$,$S[f]$, and the additive noise $N[t,f]$ are described in Sec. \ref{sec:System Model}.
The other signals in \eqref{eq:in out v1} are: (i) $Z_{SN}[t,f]$-- shot noise: modeled as a a Poisson process over each of the frequency bands. This noise is a form of noise that naturally occurs in photon counting in optical devices. 
(ii) $Z_{CN}[t,f]$-- cosmic noise: is modeled, for each frequency $f$, as a train of impulses of amplitude $A$ and inter-impulse time the impulse Poisson distributed.
%
Cosmic corresponds by electromagnetic waves generated by celestial objects such as quasars. 
Finally  $X_{FN,j}[f]$ is the fluorescence background signal emitted by the analyte $j$.
%
%
The florescence background signal is modeled as a low-degree polynomial which is a deterministic signal  in frequency but unknown at the time of the experiment. 	Florescence occurs as  Raman-active molecules also exhibit fluorescence upon excitation in conventional Raman spectroscopy

In choosing the  signal model of \eqref{eq:in out}, we assume that (i) the effects of the shot noise are also captured in the additive noise,  (ii) the time instants in which the cosmic noise occurs can be removed by hand and (iii) the data is pre-processed so at to remove the last florescence. 

The florescence background removal process for time recording of gly is shown in Fig. \ref{fig:removeflorescence}.
The florescence background is estimated from the time-averaged data through polynomial fitting with the loss function
\ea{
l(y,\yh)=  \lcb
\p{
(y-\yh)^2& y\geq \yh \\
\infty & \yh > y.
}
\rnone
\label{eq:pos mse}
}
We refer to the quantity in \eqref{eq:pos mse} as the \emph{positive MSE loss}, since  it corresponds to the MSE loss when the true data is above the estimated one but infinity otherwise. 
As a result, of the choice of loss function in \eqref{eq:pos mse}, the estimated florescence is never larger than the recorded Raman signal.

\begin{figure}
	\centering
	\begin{subfigure}{0.45\textwidth}
		\centering
		\includegraphics[trim={2cm 0 1.5cm 1.25cm},clip,width=0.85\textwidth]{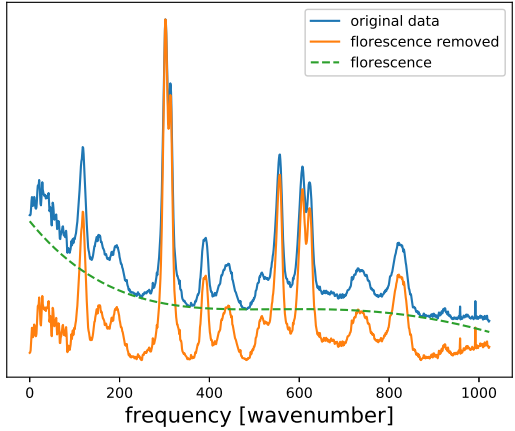}
		\caption{Florescence removal for gly as discussed is App. \ref{app:modelling}.}
		\label{fig:removeflorescence}
	\end{subfigure}
	\hfill
	\begin{subfigure}{0.48\textwidth}
		\centering
		\includegraphics[trim={2cm 0 1.5cm 1.25cm},clip,width=\textwidth]{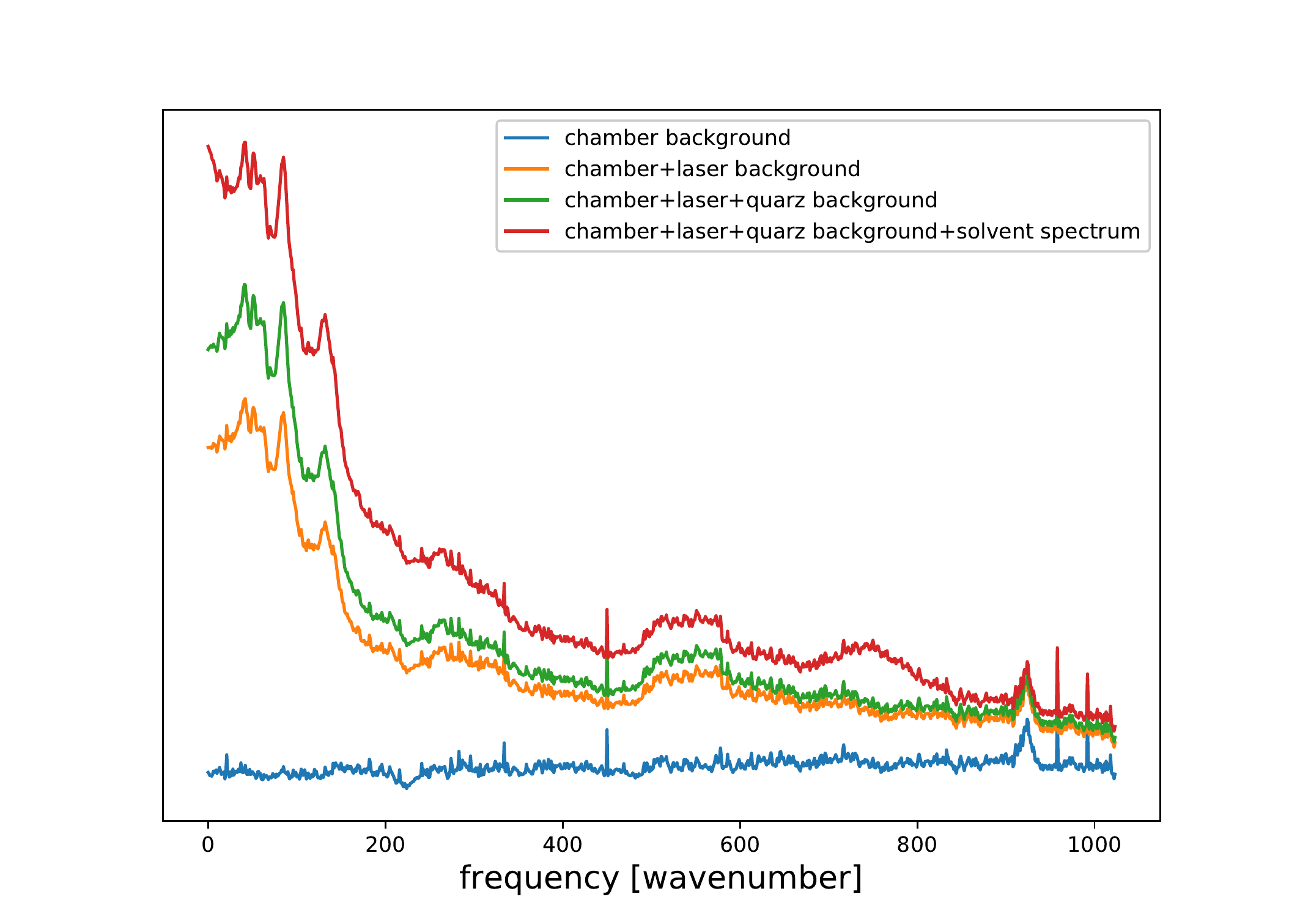}
		\caption{The backgrounds signals appearing in the data recordings, as described in App. \ref{sec:Data preparation amino}.}
		\label{fig:removeflorescence 2}
	\end{subfigure}
	\vspace{-0.15 cm}
	\caption{Further details on the data processing.}
	\label{fig:removeflorescence tot}
	\vspace{-0.5 cm}
\end{figure}

\section{Data processing in figures}

Following from  App. \ref{app:modelling}, let us further comment on the preparation of the data used in Fig. \ref{fig:amino}, Fig. \ref{fig:tuned PCA}, Fig. \ref{fig:expert} and Fig. \ref{fig:us}. 
\subsection{Data processing for Fig. \ref{fig:amino}}
\label{sec:Data preparation amino}
High concentration of the amino-acids are recorded for $20$ sec.: the  exposure of of each data point is $0.2$ sec, so that $100$ spectra are recorded.
The HPLC system (here and in subsequent data recordings) has a flow rate is $7$ ml/ min, which means $50$ uL sample takes $0.43$ sec. to flow through the vertical flow.
%

The data recorded in each session is processed so as to remove background signals arising from (i) the recording chamber (i.e. dark), (ii)  the imperfect cancellation of the laser,  and (iii) the quartz lens used focusing.
The signal arising as the superposition of the various background signal above is presented as in Fig. \ref{fig:removeflorescence2}.
In each recording, the various component of the HPLC Raman instrumentation are added, giving rise to an accumulation of the background signal. 
The last recording (red line) corresponds to  the unprocessed output of the solvent spectrum.
%

The spectra in Fig. \ref{fig:spectra} are obtained by recording high concentrations of the corresponding amino acids and removing the background signal corresponding to the dark, the laser and the quartz. 
Successively, the estimated florescence background is removed as detailed in App. \ref{app:modelling}.

The time elution in Fig. \ref{fig:time} are instead recorded through UV light absorption.

\subsection{Data processing for Fig. \ref{fig:tuned PCA}}
\label{app:tuned PCA}

The data in Fig. \ref{fig:tuned PCA} is produced from the experimental setting discussed in Sec. \ref{sec:Experimental Results} in which   $100$ mM of gly, $100$ mM of leu, $55$ mM of phe, and $55$ mM of trp  are eluted.
The PCA components are obtained after normalization through standard techniques and the matrix $T$ is carefully designed so that the reconstructed components most closely resemble the desired frequency and time characteristics. 
The time and frequency components of the four analytes are alternatively represented as outer products:  see Fig. \ref{fig:intensity_plot}.

\begin{figure}
	\centering
	\vspace{-0.5 cm}
	\includegraphics[trim={2cm 0 1.5cm 1.25cm},width=1.1\linewidth]{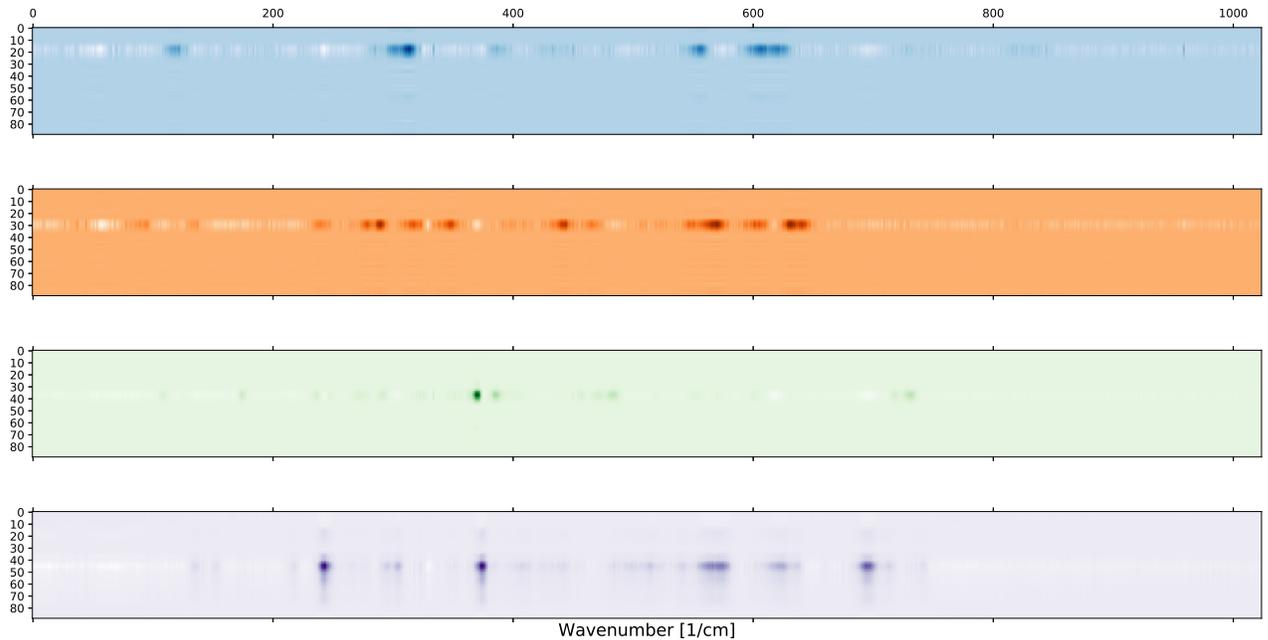}
	\vspace{-0.15 cm}
	\caption{A representation of the time-frequency components reconstructed through PCA as in Sec. \ref{app:tuned PCA}.} 
	\label{fig:intensity_plot}
\end{figure}
\bigskip

\subsection{Data processing for  Fig. \ref{fig:compare}}
\label{sec: last two figs}
Fig. \ref{fig:expert}, the analyte time and frequency features as identified in Sec. \ref{sec:Dimensionality reduction approach} are fitted through the time and frequency windows in \eqref{eq:cos win} and \eqref{eq:pseudo} respectively. 
The parameter of the reconstructed components are then used to produce the Fig. \ref{fig:expert}. 

\medskip

Let us present in some more detail the data processing in Algorithm  \ref{algo:detection} which yields Fig. \ref{fig:us}.
In Fig. \ref{fig:Solvent_subtraction} we plot the results of the subtraction of the known solvent spectrum at each time instant, as in Line 3 of  Algorithm  \ref{algo:detection}.
In Fig.\ref{fig:peak_finding} we plot the operations in  Line 4 and 5 of  Algorithm  \ref{algo:detection}: the smoothing uses Savitzky-Golay filter of order $3$ and windown length $9$, and the peak-finding algorithm using a wavelet transformation.

\begin{figure}
	\centering
	\begin{subfigure}{0.45\textwidth}
		\centering
		\includegraphics[trim={2cm 0 1.5cm 1.25cm},clip,width=\textwidth]{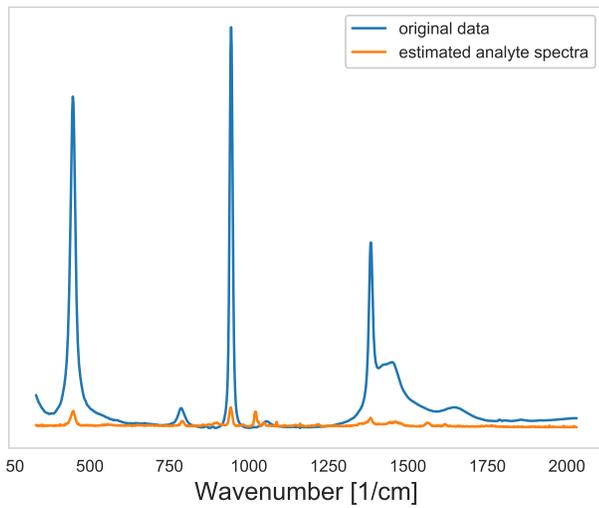}
		\caption{Solvent subtraction in Line 3 of  Algorithm  \ref{algo:detection}.}
		\label{fig:Solvent_subtraction}
	\end{subfigure}
	\hfill
	\begin{subfigure}{0.5\textwidth}
		\centering
		\includegraphics[trim={2cm 0 1.5cm 1.25cm},clip,width=\textwidth]{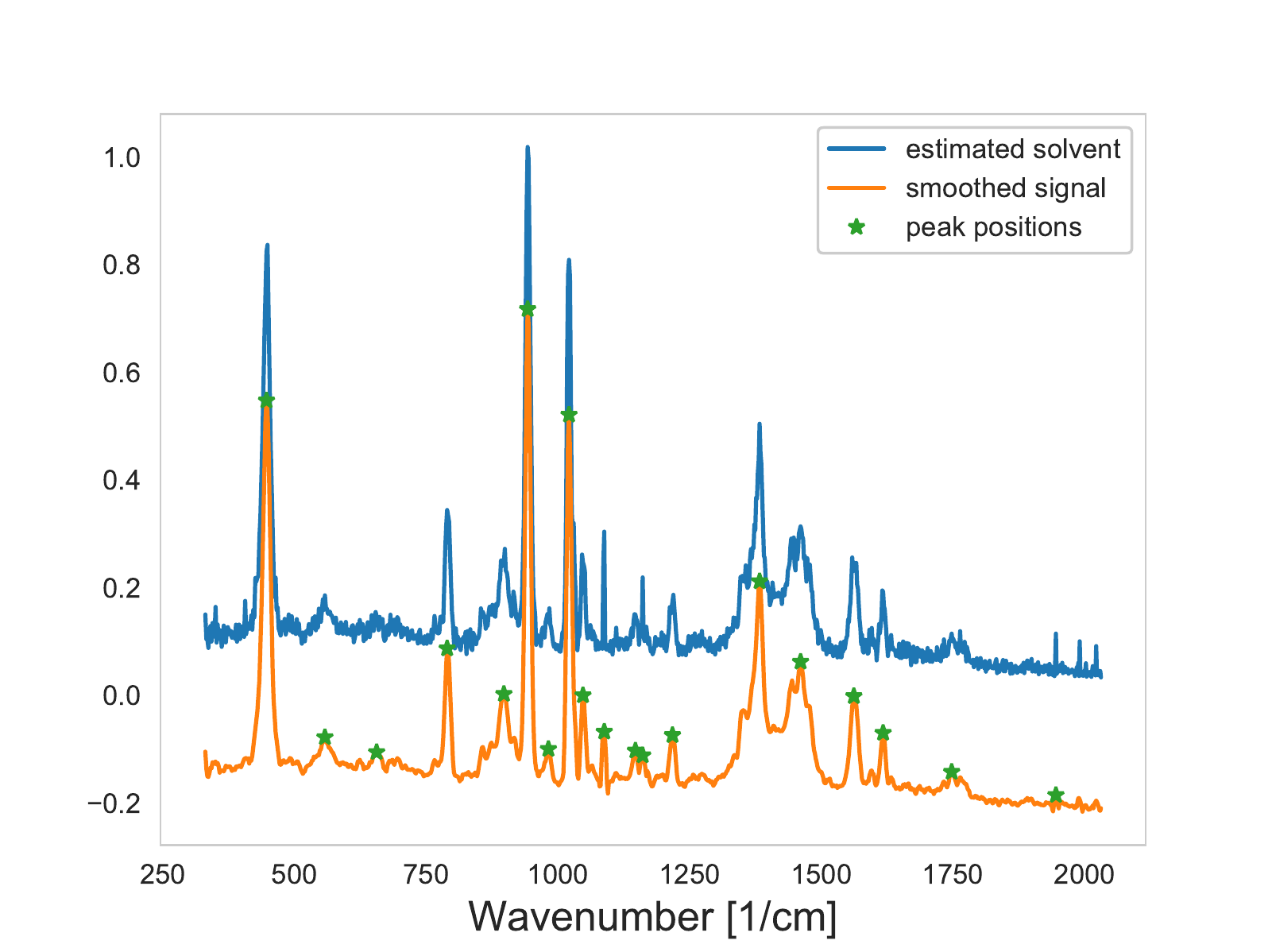}
		\caption{Smoothing and peak-finding in Line 4 and 5 of  Algorithm  \ref{algo:detection}.}
		\label{fig:peak_finding}
	\end{subfigure}
	\vspace{-0.15 cm}
	\caption{Data processing in Algorithm  \ref{algo:detection} as discussed in Sec. \ref{sec: last two figs}.}
	\label{fig:Solvent_subtraction and peak_finding}
	\vspace{-0.5 cm}
\end{figure}

In Fig. \ref{fig:fitting app}  we plot the time and frequency fitting for the gly peak at frequency $f=1020$.
The parameters of the fitting are used, together with those of the other peaks in \eqref{fig:peak_finding}, to cluster the analytes in the respective time windows using the $K$-means algorithm, as plotted in Fig. \ref{fig:us}. 
\begin{figure}[h]
	\centering
	\begin{subfigure}{0.45\textwidth}
		\centering
		\includegraphics[trim={2cm 0 1.5cm 1.25cm},clip,width=\textwidth]{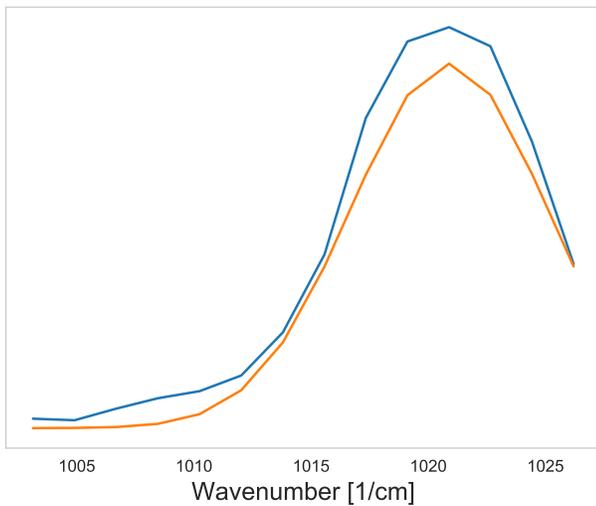}
		\caption{Gly peak spectral fitting as in App. \ref{sec: last two figs}.}
		\label{fig:frequency fitting app}
	\end{subfigure}
	\hfill
	\begin{subfigure}{0.5\textwidth}
		\centering
		\includegraphics[trim={2cm 0 1.5cm 1.25cm},clip,width=\textwidth]{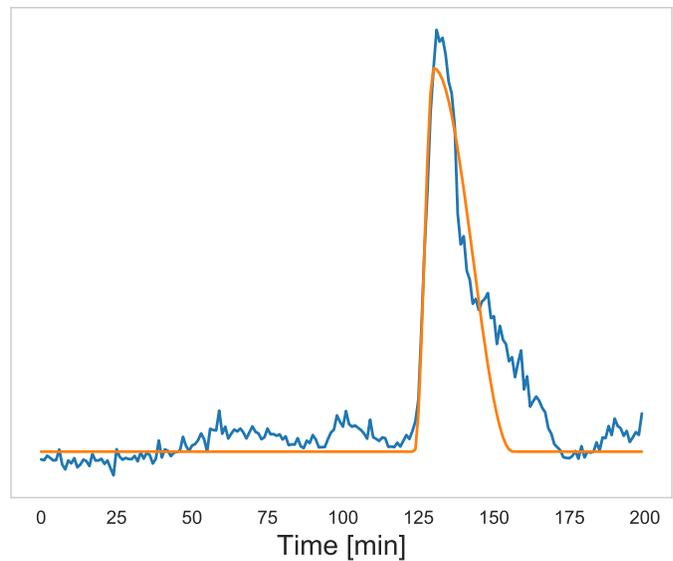}
		\caption{Gly peak temporal fitting as in App. \ref{sec: last two figs}.}
		\label{fig:time fitting app}
	\end{subfigure}
	\vspace{-0.15 cm}
	\caption{Time/frequency peak fitting for gly as described in App. \ref{sec: last two figs}. }
	\label{fig:fitting app}
	\vspace{-0.5 cm}
\end{figure}
\end{document}